\documentclass[conference]{IEEEtran}
\IEEEoverridecommandlockouts
\usepackage{cite}
\usepackage{amsmath,amssymb,amsfonts}
\usepackage{algorithmic}
\usepackage{graphicx}
\usepackage{textcomp}
\usepackage{xcolor}

\def\BibTeX{{\rm B\kern-.05em{\sc i\kern-.025em b}\kern-.08em
    T\kern-.1667em\lower.7ex\hbox{E}\kern-.125emX}}

\begin{document}

\title{Region-controlled Style Transfer\\
\thanks{Identify applicable funding agency here. If none, delete this.}
}
\author{\IEEEauthorblockN{1\textsuperscript{st} Junjie Kang}
\IEEEauthorblockA{\textit{dept. name of organization (of Aff.)} \\
\textit{name of organization (of Aff.)}\\
Guilin, China \\
2745451610@qq.com}
\and
\IEEEauthorblockN{2\textsuperscript{nd} Jinsong Wu}
\IEEEauthorblockA{\textit{dept. name of organization (of Aff.)} \\
\textit{name of organization (of Aff.)}\\
Guilin, China \\
email address or ORCID}
\and
\IEEEauthorblockN{3\textsuperscript{rd} Shiqi Jiang}
\IEEEauthorblockA{\textit{dept. name of organization (of Aff.)} \\
\textit{name of organization (of Aff.)}\\
Guilin, China \\
email address or ORCID}
\and
\includegraphics[width=\textwidth]{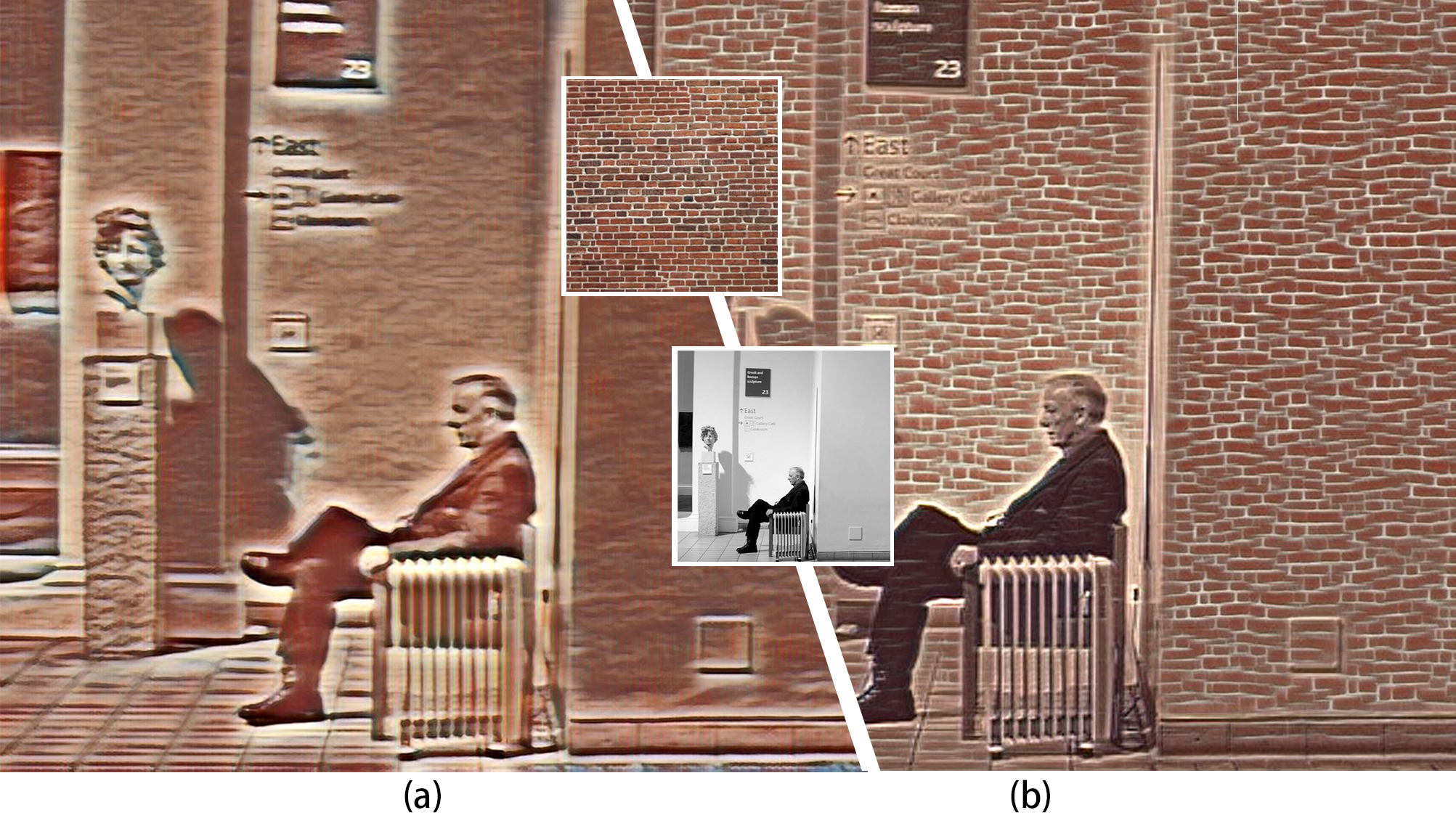}\\
\small In the given images, (a) depicts the effectiveness of Microast \cite{wang2022microast}, whereas (b) illustrates the outcomes of our proposed algorithm\\

}

\maketitle

\begin{abstract}
Image style transfer is a challenging task in computational vision. Existing algorithms transfer the color and texture of style images by controlling the neural network's 
feature layers. However, they fail to control the strength of textures in different regions of the content image. To address this issue, we propose a training method that uses a loss function to constrain the style intensity in different regions. This method guides the transfer strength of style features in different regions based on the gradient relationship between style and content images. Additionally, we introduce a novel feature fusion method that linearly transforms content features to resemble style features while preserving their semantic relationships. Extensive experiments have demonstrated the effectiveness of our proposed approach.
\end{abstract}


\begin{figure*}
  \centering
  \includegraphics[width=0.7\textwidth]{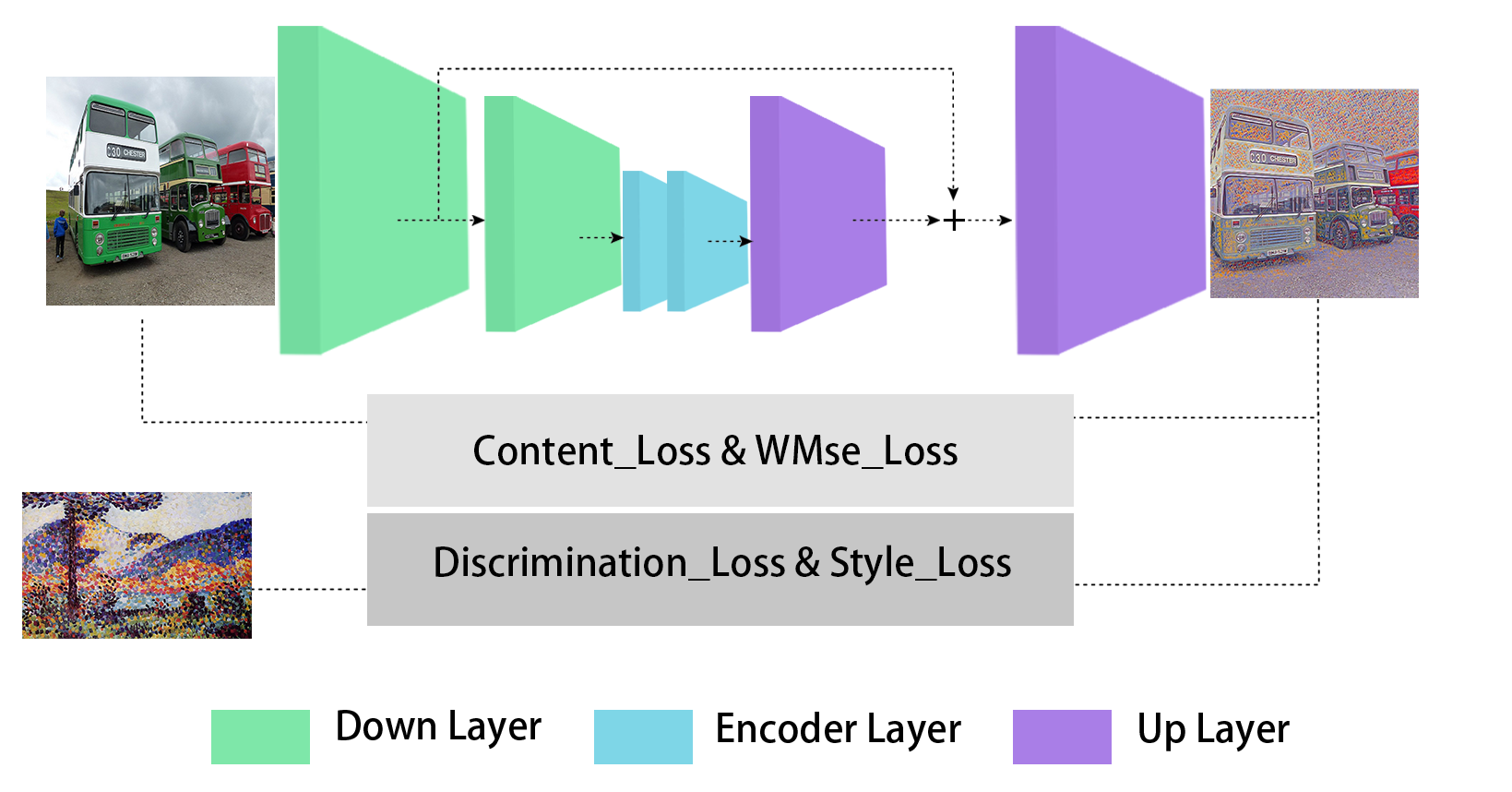} 
  \caption{Network architecture: image presentation demonstrate the overall structure of the algorithm proposed in this paper, along with the loss functions used for training the neural network. These include the style loss and content loss based on the pretrained VGG network, as well as the adversarial loss from the pretrained discriminator. Additionally, we introduce the weighted MSE loss in this paper.} 
  \label{network} 
\end{figure*}

\section{Introduction}
Image style transfer aims to extract multi-dimensional features such as color, texture, material, and emotion from style images and apply them to content images while preserving the semantic content and details of the content image. In recent years, neural network-based style transfer has rapidly evolved. \cite{ref1}introduced neural network for the first time to process this task. \cite{adain} used AdaIN normalization to linearly fuse content and style features, enabling the resulting image, after decoding, to simultaneously possess the content from the content image and the style from the style image. However, AdaIN normalization lacks the ability to interactively compute feature intensities across different channels, and due to the positive definite nature of AdaIN's transformation matrix, it cannot reverse signal variation gradients. As a result, the generated textures often appear less pronounced, and there is a heavy overlap between content and texture. \cite{stylegan2}replaced AdaIN's feature fusion method with modulation-demodulation in StyleGAN, which effectively mitigates image artifacts according to the authors. \cite{stylegan2} also introduced self-attention mechanism for feature fusion, augmenting inter-channel information on top of AdaIN and achieving superior results.\cite{our1} propose a lightweight framework that enables linearly increasing the intensity of texture transfer.\cite{our2} propose a lightweight network that can simultaneously achieve color transfer and texture transfer, achieving state-of-the-art results in both tasks.
\cite{diffusion_style_tramsfer},\cite{artfusion_style_transfer2},\cite{diffusion_style_transfer3} employed diffusion for image style transfer, offering a more diverse range of transfers that can not only replace textures but also reasonably modify the image content. However, diffusion-based methods entail significantly higher computational costs compared to other types of algorithms.

Our paper builds upon the following findings: existing methods lack the ability to control texture intensity in different regions, often suffer from color leakage from the content image, and consequently compromise the texture color obtained through transfer. As depicted in title Figure (a), the intensity of texture remains uniform in both blank regions and regions with dense details. This uniform distribution undermines the aesthetic appeal of texture transfer itself, as excessive texture in detail-rich areas can lead to the loss of content image details.

To address these limitations, we propose a novel intensity constraint loss that allows for the control of texture intensity in different regions based on their smoothness. This approach aims to preserve content appropriately, enhance the aesthetic appeal of texture transfer in different regions, and prevent color leakage from the content image. The effectiveness of our method is demonstrated in title image (b), where regions with prominent content structures exhibit reduced transferred texture information, while blank regions showcase rich patterns of texture and color.
Our contributions can be summarized as follows:
\begin{itemize}
\item We introduce a novel weighted MSE content loss that preserves only the contours and semantic information of the content image while avoiding color leakage issues by discarding continuous color information.

\item Since the style loss and content loss are inherently adversarial, the zero region of the weighted MSE content loss reduces the adversarial effect with the style loss, leading to better texture and color transfer effects.

\item Through extensive experiments, we demonstrate the effective constraint of style intensity in different regions using our proposed method on various datasets.
\end{itemize}
\begin{figure*}[htb]
  \centering
  \includegraphics[width=0.8\textwidth]{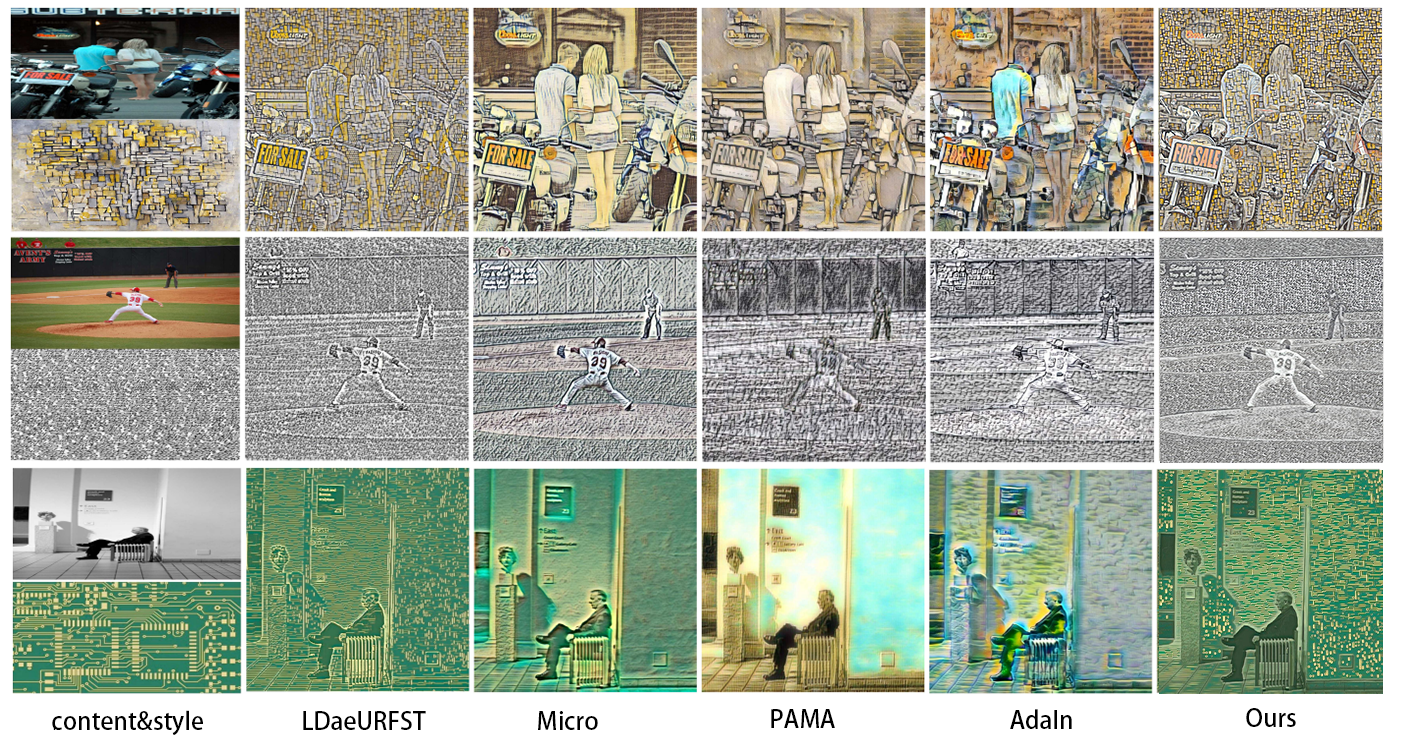} 
  \caption{The image presents the results of comparative experiments on different content and style datasets for the LDaeURFST\cite{our1}, Micro\cite{wang2022microast}, PAMA\cite{pama}, AdaIN\cite{adain}, and RCST(our proposed) algorithms.} 
  \label{experiment} 
\end{figure*}
\section{Related work}
The problem of image style transfer can be divided into sub-problems such as texture transfer and color transfer. Our focus is on extracting and transferring texture and color from style images while preserving the content of the content image. Before the era of neural networks, traditional algorithms relied on statistical measures such as histograms, mean color values, and variances to manually design algorithms for color transfer. For texture signals with more semantic features, traditional methods required separate mathematical modeling for different texture features \cite{texture1},\cite{texture2},\cite{texture3}. These approaches lacked efficiency and were not practical for real-world applications.

With the advent of neural networks, neural network-based style transfer algorithms were initially proposed by \cite{ref1}, and since then, new methods based on this framework have been continuously developed \cite{vgg_base1}, \cite{vgg_base2}, \cite{vgg_base3}, \cite{vgg_base4}, \cite{wct2}. These methods often use pre-trained VGG \cite{vgg} networks from classification tasks as feature extractors for image and content features. Researchers have found that specific layers of VGG can serve as metrics for measuring image content structure, texture, and color information. Furthermore, in existing transfer frameworks, VGG  is also incorporated as part of the loss function to constrain the convergence of the neural network.

It is well-known that the most important component of image content is the edge information, followed by color, detail, and texture features. However, this information can often be blurred or replaced by the transferred new texture. Additionally, color information should be considered as part of the texture information, and the transferred new texture should possess the relevant colors from the style image. In the existing training framework, the neural network adopts a balanced approach to balancing edge coverage and color leakage, resulting in the neural network applying the same texture intensity to different regions of the image. To address these issues, we propose a novel training approach that constrains the spatial gradients of the content image, thereby achieving the goal of differentiating texture intensities. In densely detailed regions, we can cover more texture and color features from the style image, while in smooth spatial regions, we weaken the style texture but maintain the transferred color features.

Overall, our proposed approach aims to address the challenges in extracting and transferring texture and color while preserving the content of the image.

\section{Method}
Figure \ref{network} showcases the basic network architecture we adopt, which consists of four fundamental components: an ultra-lightweight content encoder, an ultra-lightweight style encoder, a pretrained VGG network \cite{vgg}, and a lightweight adversarial discriminator \cite{gan}. These components are utilized for encoding the content structure, color and texture of the images. The basic framework has been proposed by our team to enable fast inference of specific styles on mobile devices \cite{our1}.

During the inference process, our model initially performs channel expansion and spatial dimension reduction on the input content using the initial convolutional layers. This step serves two purposes: firstly, mapping different channel information to a high-dimensional space helps cluster and decouple coupled image features, as RGB spatial information is often deeply entangled. Secondly, reducing the spatial dimensions of the input content efficiently decreases computational costs for subsequent calculations.

Next, the obtained initial features undergo dual-channel processing. One channel, referred to as the shallow feature channel, is primarily responsible for preserving the structural information of the content image. The other channel, the deep feature channel, employs a U-shaped network structure to learn the texture and color features of the specific style image. Finally, we combine the characteristics of both channels. As mentioned in \cite{our1}, employing different interpolation methods effectively controls the overall texture intensity. such as \ref{eq0}
\begin{equation}
\mathcal{F}_{merge} = \alpha \mathcal{F}_{S} + \beta \mathcal{F}_{D}
\label{eq0}
\end{equation}
We use subscripts $merge$, $S$, and $D$ to differentiate the fusion features, shallow layer features, and deep layer features.
However, this gradual texture intensity transition appears uniform in the spatial domain. In the next section, we will discuss how to address this issue by incorporating loss constraints.

\subsection{weighted mse loss}
During the training process on images with different styles and contents, we observed that the representation of texture features from style images and structural features from content images in the backbone network was inconsistent. When we iteratively passed the content image through the network for stylization, as shown in the following equation:
\begin{equation}
G(x)=G(....G(G(x)))
\label{eq1}
\end{equation}
where G means the neural network, furthermore,  the x means the input of content image.
The structural features of the content image are continuously weakened, while the texture features of the style image are progressively enhanced. However, this enhancement is not a finer-grained coverage, such as applying denser textures while preserving the content's structural edges. Instead, it simply weakens the image's structural features.

This indicates that the inference process of the network itself follows a pattern of satisfying loss requirements by superimposing texture information from the style image, indirectly diminishing the content's structural features. The explicit representation of structural features in the final output image relies on their highly responsive values in the abstract feature maps. On the other hand, texture superimposition involves a rough overlay on the feature maps, where the activated regions of the texture feature maps not only cover the unresponsive regions of the content feature maps but also overlap with the responsive regions.

Based on the above analysis, we believe that the intensity of texture feature superimposition can be controlled by applying different degrees of response constraints to different regions. In blank regions of the content image, texture coverage does not affect the content, while in regions with concentrated details, such coverage needs to be avoided.

\begin{equation}
\mathcal{L}_{mse} = ||G(x_{i+1}) - G(x_{i})||_{2}
\label{eq2}
\end{equation}

\begin{equation}
\mathcal{L}_{weighted} = \nabla x \mathcal{L}_{mse}
\label{eq3}
\end{equation}
where \eqref{eq2} represents the original $mse$ loss, and the \eqref{eq3} display the weighted $mse$ loss which our produced.$\nabla$ means operation of the Laplace with content image.

The equation \eqref{eq2}-\eqref{eq3} above represents our proposed weighted mean squared error (MSE) loss. It first calculates the gradient changes in different directions using gradient operators in the spatial domain. In regions with content edges or abundant details, there are significant gradient variations, and the gradient map has highly concentrated highlighted values. In blank regions, the gradient changes approach zero, and the gradient map has sparse highlighted values.
\begin{figure}[htb]
  \centering
  \includegraphics[width=0.35\textwidth]{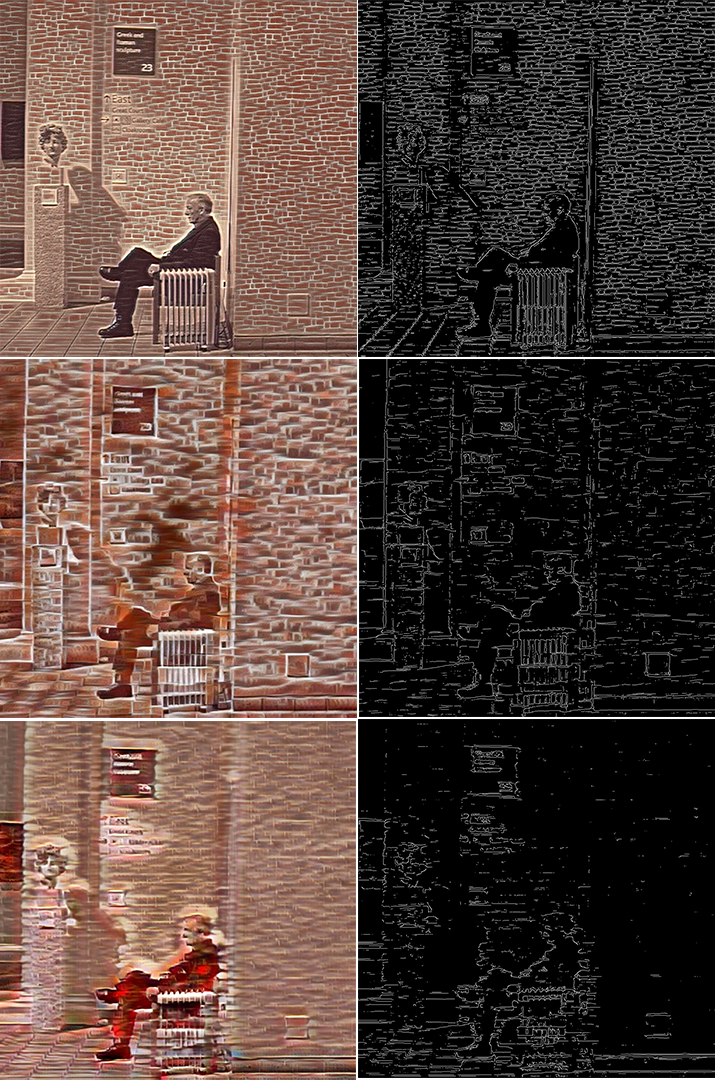} 
  \caption{The image showcases the feature detection performance of various algorithms for style transfer. From top to bottom, the sequence includes our proposed algorithm, AdaIN \cite{adain}, and SANet\cite{vgg_base4}.} 
  \label{edge} 
\end{figure}
\begin{figure*}[htb]
  \centering
  \includegraphics[width=0.8\textwidth]{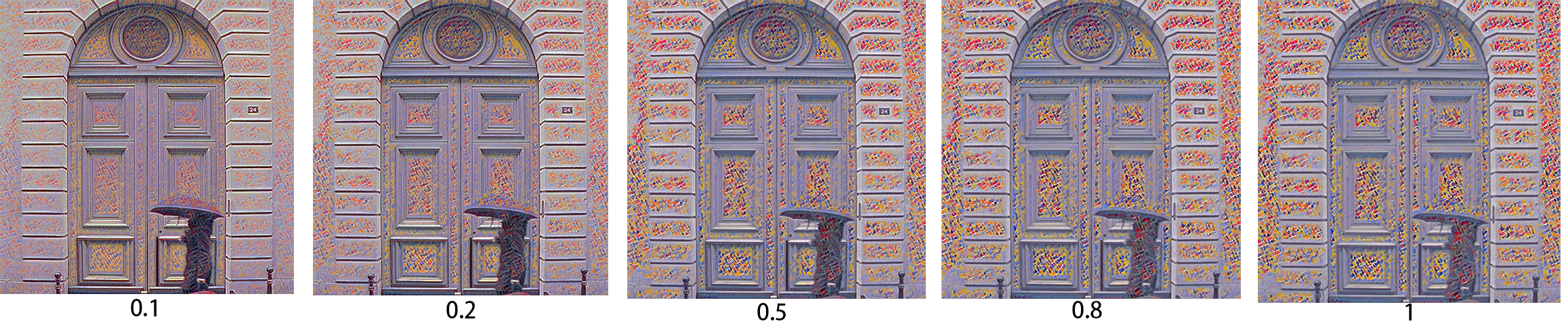} 
  \caption{The image illustrates the results obtained by combining deep features and shallow features with different coefficients.} 
  \label{jianbian} 
\end{figure*}
The gradient change map can be used as the weight for the MSE loss in different regions. Thus, details preservation is neglected in blank regions, while it is emphasized in regions with details and edges.

\subsection{region-wise controllable interpolation}
The final computation result requires the evaluation of content features from shallow channels, color features, and texture features from deep channels. In the first subsection of this chapter, we mentioned that these two sets of features can assign different texture intensities to the output image by weighting them with different coefficients \ref{eq0}.Typically, the coefficients controlling shallow features are set to $1$. With the inclusion of the weighted MSE loss\ref{eq3}, the variation in texture intensity only affects non-structural regions and does not compromise the details of the content structure during the process of gradual enhancement. We will demonstrate this effect during the experimental phase.

\section{experiment}
Our experiments were conducted on an RTX 3090, using the COCO dataset \cite{cocodata} as the training set. Remarkably good results can be achieved with just 6 to 8 epochs of training. 

\subsection{comparative experiments}

Firstly, we qualitatively compared the results of various style transfer algorithms, such as \cite{adain},\cite{wang2022microast},\cite{pama}, \cite{our1}, \cite{}, under different content and style combinations. As shown in Figure \ref{experiment}, our proposed algorithm demonstrates strong regional characteristics in the transferred textures. In areas with structured content, such as people and buildings, the transferred textures are appropriately suppressed, preserving the structural details of the content image. On the other hand, in blank areas, there is a distinct texture effect. When compared to other algorithms, such as [1][2], the transferred textures and colors are not as prominent, or algorithms like [3][4] exhibit the same texture density across different structural regions.

In addition, to demonstrate the preservation of edge structures and details in the content images by our proposed algorithm, we conducted a comparison of edge preservation among different algorithms. As shown in Figure \ref{edge}, our algorithm exhibits clearer texture edges as well as highly defined edges of the content structure. The transferred texture does not compromise the inherent content structure. In contrast, the compared algorithms show less distinct texture edges overall and disrupt the edges of the content. It is evident that our algorithm exhibits stronger edge preservation capabilities.

Furthermore, due to the inherent interpolation properties of our algorithm, we will showcase the control of style transfer intensity with different feature coefficients. As shown in Figure \ref{jianbian}, our algorithm maintains the contrast in style intensity between content and non-content regions under various feature coefficients. Particularly, at higher feature coefficients, our algorithm demonstrates distinct regional separation characteristics. It is important to note that all of these results were achieved without the use of semantic segmentation.

\subsection{ablation experiment}
\begin{figure}[htb]
  \centering
  \includegraphics[width=0.42\textwidth]{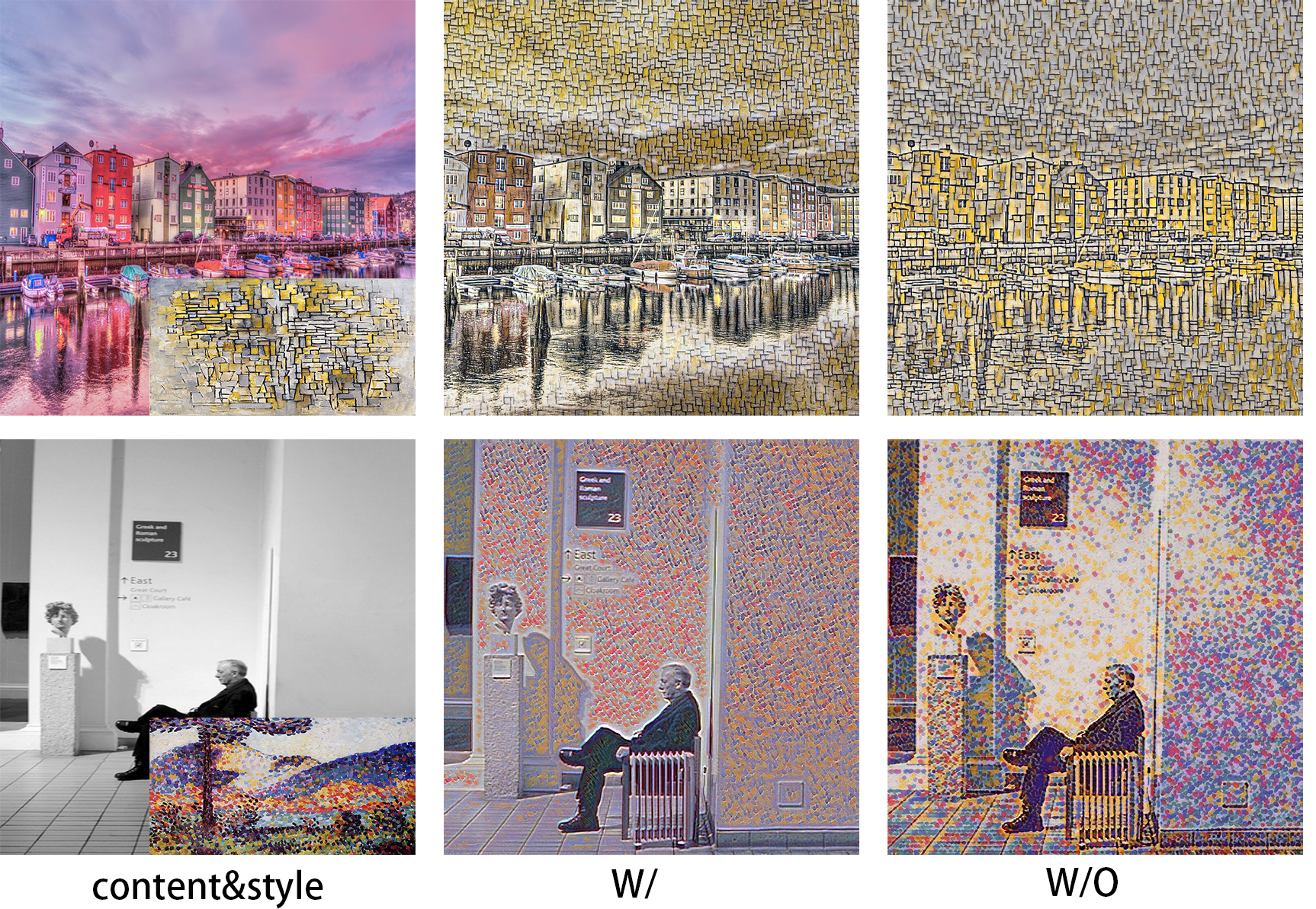} 
  \caption{The image showcases the results of ablation experiments. Each row, from left to right, consists of the content image with style image, results obtained using weighted MSE training, and results obtained without weighted MSE training.} 
  \label{ablation} 
\end{figure}

In this section, we will demonstrate the effectiveness of our proposed loss function. Figure \ref{ablation} presents the results of ablation experiments. The left image shows the results obtained using weighted MSE training, while the right image shows the results obtained without weighted MSE training. It is evident that with the constraint of gradient weights, the structural features of the images are better preserved. Moreover, overall, our algorithm does not compromise the transfer results in other regions.
\section*{conclusion}
Our algorithm provides a partial elucidation of the texture generation mechanism in the context of style transfer and addresses the issue of spatial averaging in texture transfer regions. We propose a weighted variance loss to maintain consistency in content structure across different model pathways. Our algorithm is concise yet effective, as we avoid the use of techniques like segmentation and instead rely on gradient-based constraints. Extensive experimentation confirms its effectiveness. However, there is still room for improvement in our algorithm. Firstly, we aim to generalize the framework from style-specific to arbitrary style transfer. Secondly, we seek to enhance the speed and efficiency of inference. While our algorithm exhibits minimal computational overhead, the computational cost grows exponentially with larger input images, presenting a critical challenge. Finally, we plan to incorporate additional control mechanisms into our framework, such as multimodal semantic control, to generate advanced textures using language models such as CLIP \cite{clip}. These improvement suggestions will guide our future research endeavors.

\bibliographystyle{plain}
\bibliography{ref}

\end{document}